%% file: 282_main_paper.tex
\documentclass[letterpaper]{article}
\usepackage{uai2020}
\usepackage[margin=1in]{geometry}

\usepackage[sort&compress,numbers]{natbib}

\usepackage{times}

\usepackage[autonum]{tchdr}
\usepackage{tikz}
\usetikzlibrary{bayesnet}

\newcommand{\Kexp}{\kappa^\text{exp}}

\title{Slice Sampling for General Completely Random Measures}

\author{ 
{\bf{Peiyuan Zhu} \qquad \bf {Alexandre Bouchard-C\^{o}t\'{e}} \qquad \bf {Trevor Campbell}}  \\
Department of Statistics \\
University of British Columbia\\
Vancouver, BC V6T 1Z4 \\
}

\begin{document}

\maketitle

\input{abstract}

\input{introduction}
\input{background}

\input{slicesampling}

\input{applications}

\input{experiments}

\input{conclusion}

\input{acknowledgement}

\appendix
\input{282_supplementary}

{
	\small
	\bibliographystyle{unsrtnat}
	\bibliography{sources}
}

\end{document}

%% file: abstract.tex
\begin{abstract} 
Completely random measures provide a principled approach to
creating flexible unsupervised models, where the number of latent features is
infinite and the number of features that influence the data grows 
with the size of the 
data set.  Due to the infinity the latent features, posterior inference
requires either marginalization---resulting in dependence structures that
prevent efficient computation via parallelization and conjugacy---or finite 
truncation, which arbitrarily limits the flexibility of the model.  In
this paper we present a novel Markov chain Monte Carlo algorithm for posterior
inference that adaptively sets the truncation level using auxiliary slice
variables, enabling efficient, parallelized computation without sacrificing
flexibility.  In contrast to past work that achieved this on a model-by-model
basis, we provide a general recipe that is applicable to the broad class of
completely random measure-based priors.  The efficacy of the proposed algorithm
is evaluated on several popular nonparametric models,
demonstrating a higher effective sample size per second compared to algorithms using
marginalization as well as a higher predictive performance compared to models
employing fixed truncations.
\end{abstract}

%% file: introduction.tex
\section{INTRODUCTION}

In unsupervised data analysis, one aims to uncover complex latent structure in data.
Traditionally, this structure has been assumed to take the form of a 
\emph{clustering}, in which each data point
is associated with exactly one latent category.  Here we are concerned with a new
generation of unobserved structures such that each data point can be associated
to any number of latent categories.  When such model can select each category
zero or once, the latent categories are called \emph{features}
\cite{griffiths05}, whereas if each category can be selected with
multiplicities, the latent categories are called \emph{traits}
\cite{campbell18}.  

Consider for example the problem of modelling movie ratings for a set of users.
As a first rough approximation, an analyst may entertain a clustering over the
movies and hope to automatically infer movie genres.  Clustering in this
context is limited; users may like or dislike movies based on
many overlapping factors such as genre, actor and score preferences.  Feature
models, in contrast, support inference of these overlapping movie attributes. 

As the amount of data increases, one may hope to capture increasingly sophisticated
patterns. We therefore want the model to increase its complexity accordingly.
In our movie example, this means uncovering more and more diverse user
preference patterns and movie attributes from the growing number of registered
users and new movie releases. Bayesian nonparametric methods (BNP) enable
unbounded model capacity by positing infinite-dimensional prior
distributions. These infinite dimensional priors are designed so that for any
given dataset only a finite number of latent parameters are utilized, making
Bayesian nonparametric inference possible in practice.
The present work is concerned with developing efficient and flexible inference
methods for a class of BNP priors called completely random measures (CRMs) \cite{kingman67},
which are commonly used in practice \cite{cai16,gupta13,paisley12,acharya15}. In 
particular, CRMs
provide a unified approach to the construction of BNP priors over both latent features 
and traits \cite{jordan10,broderick18}. 

Previous approaches to CRM posterior inference can be categorized
into two main types.  First, some methods analytically marginalize the infinite
dimensional objects involved in 
CRMs \cite{griffiths05,thibaux07,titsias08,griffin11,doshivelez09acc,zhou12}. This
 has the disadvantage of making restrictive conjugacy
assumptions and is moreover not amenable to parallelization. A second type of
inference method introduced by \citet{blei06} instead uses a fixed truncation 
of the infinite model. However,
this strategy is at odds with the motivation behind BNP, namely, its
ability to learn model capacity as part of the inferential procedure.
\citet{campbell19} provide \emph{a priori} error bounds on such truncation, but
it is not obvious how to extend these bounds to approximation errors on the
posterior distribution. 

Our method is based on slice sampling, a family of Markov chain Monte Carlo methods first used
in a BNP context by \cite{walker07}. Slice samplers have advantages over both
marginalization and truncation techniques: they do not require conjugacy, enable
parallelization, and target the exact nonparametric posterior distribution. 
But while there is a rich literature on sampling methods for BNP latent feature and trait models---e.g., 
the Indian buffet / beta-Bernoulli process \citep{teh07,doshivelez09acc},
hierarchies thereof \citep{thibaux07},
normalized CRMs \citep{favaro13,griffin11},
beta-negative binomial process \citep{broderick15,zhou12},
generalized gamma process \citep{ayed19}, 
gamma-Poisson process \citep{titsias08}, and more---these
have often been developed on a model-by-model basis.

In contrast to these past model-specific techniques, we develop our sampler based on a
\emph{series representation} of the L\'{e}vy process \cite{rosinski01} underlying the CRM. 
In a fashion similar to \cite{kalli11}
we introduce auxiliary variables that adaptively truncate the series
representation; only finitely many latent features are
updated in each Gibbs sweep. The representation that we utilize factorizes the
weights of CRMs into a transformed Poisson process with independent and
identically distributed marks, thereby turning the sampling problem into evaluating
the mean measure of a marginalized Poisson point process over a zero-set. 

The remainder of the paper is organized as follows: \cref{sec:background} introduces the general model
that we consider, series representations of CRMs, and posterior inference
via marginalization and truncation.
\cref{sec:slicesampling} discusses our main contributions, including
model augmentation and slice sampling. \cref{sec:application} demonstrates
how the methodology can be applied to two popular latent feature models.
 Finally, in \cref{sec:experiments} we
compare our method against several state-of-the-art samplers
for these models on both real and synthetic datasets.

%% file: background.tex
\section{BACKGROUND}\label{sec:background}

\subsection{MODEL}
In the standard Bayesian nonparametric latent trait model \cite{campbell19}, we are
given a data set of random observations $(Y_n)_{n=1}^N$ generated using an
infinite collection of latent traits $(\psi_k)_{k=1}^\infty$, $\psi_k \in \Psi$
with corresponding rates $(\theta_k)_{k=1}^\infty$, $\theta_k \in \reals_+
\defined [0, \infty)$. We assume each data point $Y_n$, $n\in[N]\defined\{1, \dots, N\}$ is influenced by each
trait $\psi_k$ in an amount corresponding to an integer count $X_{nk} \in
\nats_0 \defined \{0,1,2,\dots\}$ via
\[
X_{nk} &\distind h(\cdot ; \theta_k) && n, k \in [N]\times\nats\\
Y_n &\distind f\left(\cdot ; \sum_{k=1}^\infty X_{nk}\delta_{\psi_k}\right) && n \in [N],
\]
where $\delta_{(\cdot)}$ denotes a Dirac delta measure, $h$ is a distribution on
$\nats_0$, and $f$ is a distribution on the space of observations. Note that
each data point $y_n$ is influenced only by those traits $\psi_k$ for which
$x_{nk} > 0$, and the value of $x_{nk}$ denotes the amount of influence. 

To generate the infinite collection of $(\psi_k, \theta_k)$ pairs, we use a
Poisson point process \cite{kingman92} on the product space of
traits and rates $\Psi \times \reals_+$ with $\sigma$-finite mean measure
$\mu$,
\[
\left\{\psi_k, \theta_k\right\}_{k=1}^\infty \dist \distPP(\mu) && \mu(\Psi\times\reals_+) = \infty.\label{eq:pp}
\]
Equivalently, this process can be formulated as a \emph{completely random
measure} (CRM) \cite{kingman67}  on the space of traits $\Psi$ by placing a Dirac measure
at each $\psi_k$ with weight $\theta_k$\footnote{More generally, CRMs are the
sum of a deterministic measure, an atomic measure with fixed atom locations,
and a Poisson point process-based measure as in \cref{eq:crm}. In BNP models,
there is typically no deterministic component, and the fixed-location atomic
component has finitely many atoms, posing no challenge in posterior inference.
Thus we focus only on the infinite Poisson point process-based component in
this paper.},
\[
\sum_{k=1}^\infty \theta_k \delta_{\psi_k} &\dist \distCRM(\mu). \label{eq:crm}
\]
In Bayesian nonparametric modelling, the traits are typically generated independently of the rates, i.e.,
\[
\mu\left(\dee\theta,\dee\psi\right)=\nu\left(\dee\theta\right)H\left(\dee\psi\right),
\]
where $H$ is a probability measure on $\Psi$, and $\nu$ is a $\sigma$-finite measure on $\reals_+$.
In order to guarantee that the CRM has infinitely many atoms, we require that $\nu$
satisfies
\[
	\nu\left(\reals_+\right)=\infty,
\]
and in order to guarantee that each observation $y_n$ is only influenced
by finitely many traits $\psi_k$ having $x_{nk} \neq 0$ \as, we require that 
\[
\hspace{-.2cm}\EE\left(\sum_{k=1}^\infty\ind\{X_{nk}\ne0\}\right) &=\int(1-h(0|\theta))\nu\left(d\theta\right)<\infty.\hspace{-.1cm}
\]
 To summarize, the model we consider in this paper is:
\[
\sum_{k=1}^\infty&\theta_k\delta_{\psi_k}\dist\distCRM\left(\nu\times H\right) && \\
X_{nk} &\distind h(\cdot ; \theta_k) && n, k \in [N]\times\nats\\
Y_n &\distind f\left(\cdot ; \sum_k X_{nk}\delta_{\psi_k}\right) && n \in [N].
\label{eq:model}
\]

\subsection{SEQUENTIAL REPRESENTATION}
While the specification of $\{\psi_k, \theta_k\}_k$ as a Poisson point process
is mathematically elegant, it does not lend itself immediately to 
computation. For this purpose---since there are infinitely many atoms---we 
require a way of generating them one-by-one in a sequence 
using familiar finite-dimensional distributions;
this is known as a \emph{sequential representation} of the CRM. 
While there are many such representations (see \cite{campbell19} for an overview),
here we will employ the general class of \emph{series representations},
which simulate the traits $\psi_k$ and rates $\theta_k$ via 
\[
E_j &\distiid \distExp(1) & \Gamma_k &= \sum_{j=1}^k E_j & V_k &\distiid G\\
\theta_k &= \tau(V_k, \Gamma_k) & \psi_k &\distiid H,
\label{eq:seriesrep}
\]
where $\Gamma_k$ are the ordered jumps of a homogeneous, unit-rate Poisson process on $\reals_+$,
$G$ is a probability distribution on $\reals_+$, and $\tau : \reals_+\times\reals_+ \to \reals_+$
is a nonnegative measurable function such that $\lim_{u\to\infty} \tau(v, u) = 0$ for $G$-almost 
every $v$. For each mean measure $\mu$ in \cref{eq:pp}, there are many choices 
of $G$ and $\tau$ that together yield a valid series representation for $\distPP(\mu)$,
such as the inverse-L\'evy representation \cite{ferguson72}, Bondesson representation \cite{bondesson82},
rejection representation \cite{rosinski01}, etc.

\subsection{POSTERIOR INFERENCE}

Posterior inference in the BNP model \cref{eq:model} is complicated by the
presence of infinitely many traits $\psi_k$ and rates $\theta_k$, as the
application of traditional MCMC and variational procedures would require
infinite computation and memory resources. Past work has handled this issue in
two ways: \emph{marginalization} and \emph{fixed truncation}. 

\paragraph{Marginalization} In a wide variety of CRM-based models, it is possible
to analytically integrate out the latent rates and traits \cite{broderick18},
thus expressing the model in terms of only the \iid traits $\psi_k$ and
sequence of conditional distributions for the assignments $X_n$, 
\[
\psi_k &\distiid H && k\in\nats\\
X_{n} &\dist \Pr\left(X_n = \cdot \given X_{1:n-1}\right) && n \in [N]\\
Y_n &\distind f\left(\cdot ; \sum_k X_{nk}\delta_{\psi_k}\right) && n \in [N].
\label{eq:marginalized_model}
\]
Using the exchangeability of the sequence $(X_n)_{n=1}^N$, Gibbs sampling \cite{tanner87} 
algorithms can be derived that alternate between sampling $X_n$ for each $n\in[N]$,
and sampling $\psi_k$ for each of the (finitely many) ``active traits'' $k$ such that $\sum_{n} X_{nk} > 0$.
However, because each $X_n$ must be sampled conditioned on $X_{-n}$, these methods cannot
be parallelized across $n$, making them computationally expensive with large amounts of data. 

\paragraph{Fixed truncation}  Another option for posterior inference is to truncate a sequential
representation of the CRM such that it generates finitely many traits, i.e.,
\[
 &\left(\psi_k, V_k, \Gamma_k\right)_{k=1}^K \dist \text{\cref{eq:seriesrep}}\\
X_{nk} &\distind h(\cdot ; \tau(V_k, \Gamma_k)) &&\hspace{-.4cm} n, k \in [N]\times [K]\\
Y_n &\distind f\left(\cdot ; \sum_{k=1}^K X_{nk}\delta_{\psi_k}\right) && n \in [N].
\label{eq:truncated_model}
\]
Because there are only finitely many traits and rates in this model, it is not difficult
to develop Gibbs sampling and variational algorithms \cite{roychowdhury15} that iterate between 
updating the rates $(\theta_k)_{k=1}^K$,
the traits $(\psi_k)_{k=1}^K$, and then the assignments $(X_n)_{n=1}^N$. Further, the 
independence of the assignments across observations $n$ conditioned on the rates and traits
enables computationally efficient paralellization of the $X$ update. 
However, the major drawback of this approach is that the error incurred by truncation
is unknown; previous work  provides bounds on the total variation
distance between the truncated and infinite data 
marginal distributions \cite{campbell19,Arbel17,argiento16,doshivelez09var}, but 
error incurred in the posterior distribution is unknown.

%% file: slicesampling.tex
\section{SLICE SAMPLING FOR CRMs}\label{sec:slicesampling}
In this section, we employ an \emph{adaptive} truncation of general CRM series
representations to obtain both the computational efficiency
of truncated methods and the statistical correctness of 
approaches based on marginalization. 
In \cref{sec:augmentedmodel} we first add an auxiliary variable
for each observation $n$ that truncates the full conditional distribution of 
its underlying assignments $X_n$.
\cref{sec:gibbssampling} provides a slice sampling scheme
for the augmented model, resulting in truncation
that adapts from iteration to iteration.

\subsection{AUGMENTED MODEL}\label{sec:augmentedmodel}
We begin by augmenting the model \cref{eq:model} with auxiliary variables $(U_n)_{n=1}^N$
that truncate the full conditional distributions of the assignments $X_n$.
In particular, suppose we fix the assignments for observations other than $n$ (denoted $X_{-n}$),
the CRM variables $\psi, V, \Gamma$, and the auxiliary variables $U$. Then we require
for some $T < \infty$,
\[
\forall k > T, \Pr\left(X_{nk} > 0 \given X_{-n}, \psi, V, \Gamma, U\right) = 0.\label{eq:truncation}
\]
Past model augmentations in Bayesian nonparametrics have
largely required either the normalization of the random measure \cite{walker07,favaro13,griffin11}
or a particular sequential representation that guarantees strictly decreasing values of $\theta_k$ \cite{teh07}.
In the present setting of general, unnormalized CRMs, we cannot take advantage of either of these facts. 

We therefore take an approach inspired by \cite{kalli11}
for augmenting the model. In particular,  
for each observation $n\in[N]$, define its maximum \emph{active index}
\[
k_n \defined \max \,\,\{k \in\nats : X_{nk}>0\} \cup \{0\},\label{eq:Kn}
\] % ABC: fixed min to max
and let $\xi : \nats_0 \to \reals_+$ be a monotone decreasing sequence
such that $\lim_{n\to\infty} \xi(n) = 0$. Then we add a uniform random
\emph{slice variable} $U_n$ lying in the interval $[0, \xi(k_n)]$ to the model 
for each observation $n$, i.e.,
\[
\hspace{-.4cm}\left(\psi_k, V_k, \Gamma_k\right)_{k=1}^\infty &\dist \text{\cref{eq:seriesrep}}\\
X_{nk} &\distind h(\cdot ; \tau(V_k,\Gamma_k)) &&\hspace{-.4cm} n, k \in [N]\times \nats\\
U_n &\distind \distUnif\left[0, \xi(k_n)\right] && n\in[N]\\
Y_n &\distind f\left(\cdot ; \sum_{k=1}^K X_{nk}\delta_{\psi_k}\right) && n \in [N].
\label{eq:adaptive_truncated_model}
\]
The variables $(U_n)_{n=1}^N$ do not change the posterior marginal of interest
on $X, \psi, \theta$, but do provide computational benefits.
In particular, the full conditional distribution of $X_n$ based on \cref{eq:adaptive_truncated_model}
sets $X_{nk} = 0$ for any $k$ such that $\xi(k) < U_n$. 
Thus, the truncation level for each $X_n$ will adapt as $U_n$ changes from iteration-to-iteration. 
Further, slice sampling in \cref{eq:adaptive_truncated_model} requires only finite memory and computation,
since we need to store and simulate only those finitely many $\psi_k, V_k, \Gamma_k$ such that $\xi(k) \geq \min_n U_n$ at 
each iteration. The ability to instantiate the latent $\psi_k, V_k, \Gamma_k$ variables
has many advantages; e.g., we can leverage the independence 
of $X_n$ for parallelization without sacrificing 
the fidelity of the model. The augmented probablistic model is depicted in \cref{fig:graph}.
\input{graph}

\subsection{SLICE SAMPLING}\label{sec:gibbssampling}
In this section, we develop a slice sampling scheme for the 
augmented model \cref{eq:adaptive_truncated_model} that iteratively
simulates from each full conditional distribution. The state
of the Markov chain that we construct is infinite-dimensional, consisting of
$(X_{nk})_{n\in [N], k\in\nats}$, $(\psi_k, V_k, \Gamma_k)_{k\in\nats}$,
and $(U_n)_{n\in[N]}$. Due to the augmentation in \cref{eq:adaptive_truncated_model}, however, only finitely
many of these variables need to be stored or simulated during any iteration
of the algorithm. The particular steps follow;
note that the order of the steps is important.

\paragraph{Initialization} Set the assignment variables $X = 0$, 
the global truncation levels $K = K_{\text{prev}} = 0$, 
and for all $n\in[N]$, the local truncation levels $k_n = k'_n = 0$. Run this step only a single
time at the beginning of the algorithm.

\paragraph{Sample $U$:} For $n\in [N]$,
draw $U_n \distind \distUnif\left[0, \xi(k_n)\right]$.

\begin{figure}[t]
\includegraphics[scale=0.33]{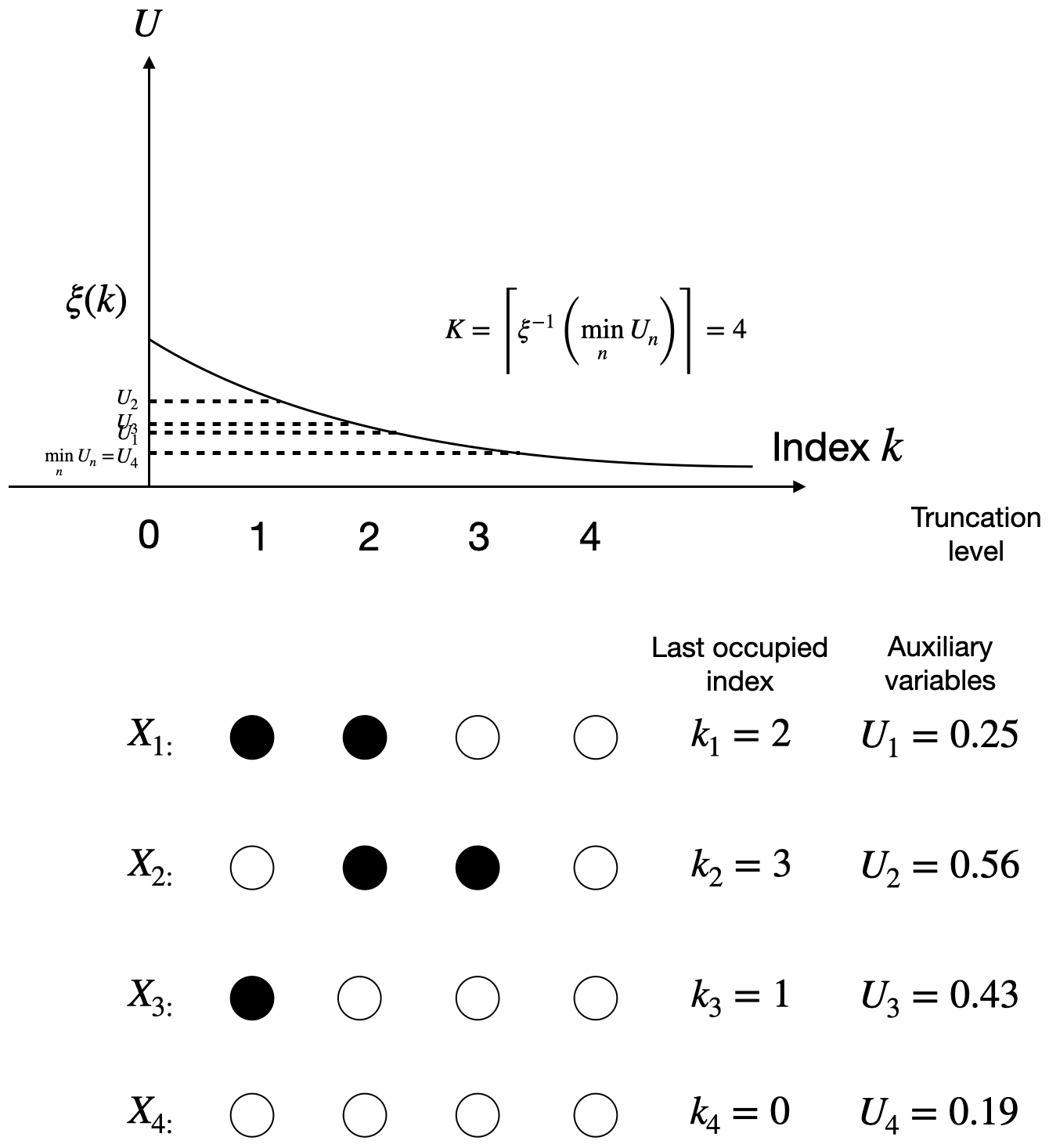}
\caption{An instance of slice variables in a dataset with four observations. }
\end{figure}{\tiny }

\paragraph{Update global truncation level:}
Set
\[
K_{\text{prev}} &\gets \max_{n\in[N]} k_n\\
K &\gets \max \left\{ k \in \nats_0 : \xi(k) \geq \min_n U_n\right\}.
\]
In other words, $K$ is the maximum index that observations might
activate in this iteration,  and $K_{\text{prev}}$ 
is the maximum index that observations activated in the previous iteration.
Note that $\Gamma_k$, $V_k$, $\psi_k$ are all guaranteed to be instantiated
for $k=1, \dots, K_{\text{prev}}$.

\paragraph{Sample $\psi$} For each $k \in [K]$, 
sample $\psi_k$ from its full conditional distribution,
with measure proportional to
\[
H(\dee\psi_k)\prod_{n=1}^Nf\left(Y_n; X_{nk}\delta_{\psi_k} +\! \sum_{j\neq k} X_{nj}\delta_{\psi_j}\right).\label{eq:samplepsi}
\]
If possible, $(\psi_k)_{k=1}^K$ should instead be sampled jointly from the same
density above. 
Further, if $f$ is not conjugate to the prior measure $H$, this step may be conducted
using Metropolis-Hastings. Since the remaining values
$(\psi_k)_{k=K+1}^\infty$ will not influence the remainder of this iteration, they
do not need to be simulated.

\paragraph{Sample $V$, $\Gamma$} This step is split into two substeps: 
first sample $(V_k, \Gamma_k)_{k=1}^{K_{\text{prev}}-1}$ each from their own full conditional,
and then sample $(V_k, \Gamma_k)_{k=K_{\text{prev}}}^\infty$ as a single block. 
Define $\Gamma_0 \defined 0$ for notational convenience.

\textbf{Substep 1:} Note that $V_k, \Gamma_k$ are conditionally independent
of all other variables given $(X_{nk})_{n=1}^N$, $\Gamma_{k-1}$, and $\Gamma_{k+1}$.
Thus, for each $k \in [K_\text{prev}]$, we generate each pair of $V_k, \Gamma_k$ from 
its full conditional, with measure proportional to
\[
G(\dee V_k)\dee\Gamma_k&\ind\left[\Gamma_{k-1}\!\le\!\Gamma_k\!\le\!\Gamma_{k+1}\right]\cdot\\
&\quad\prod_{n=1}^N h(X_{nk} ; \tau(V_k,\Gamma_k)).\label{eq:sampgvfull}
\]
\textbf{Substep 2:} If $K_\text{prev} > K$, this step is skipped.
Otherwise, for each $k \in \{K_\text{prev}, \dots, K\}$ in increasing
order, we generate $V_k, \Gamma_k$ conditioned on $V_{1:k-1}, \Gamma_{1:k-1}$
and the remaining variables. In particular, note that
$V_k, \Gamma_k$ are conditionally independent of all other variables given
$(X_{nj})_{n\in[N], j\geq k}$ and $\Gamma_{k-1}$. Thus,
we generate $V_k, \Gamma_k$ from a measure proportional to
\[
&G(\dee V_k)\dee\Gamma_k\exp\left(-\Gamma_k\right)\ind\left[\Gamma_k\geq \Gamma_{k-1}\right]\cdot\\
&\quad\Pr\left(\left(X_{nj}\right)_{n\in[N],j>k} = 0|\Gamma_k\right)\cdot\\
&\qquad\prod_{n=1}^Nh\left(X_{nk};\tau(V_k,\Gamma_k)\right)\label{eq:sampgvexp}
\]
Again, since the remaining values
$(V_k, \Gamma_k)_{k=K+1}^\infty$ will not influence the remainder of this iteration, they
do not need to be simulated and can be safely ignored.

In order to evaluate $\Pr\left(\left(X_{nj}\right)_{n\in[N],j>k} = 0\given\Gamma_k\right)$,
we use the machinery of Poisson point processes. In particular, note that 
the independent generation of $X_{nk}$ given $V_k, \Gamma_k$ ensures that
$\Pi \defined \{\Gamma_j,V_j,X_{nj}\}_{n\in[N],j>k}$ conditioned on $\Gamma_k$ 
is itself a Poisson point process on the joint space by 
repeated use of the marking theorem \cite[Ch.~5.2]{kingman92}. 
Therefore, this probability can be written as the probability that
$\Pi$ has no atoms with a nonzero integer component:
\[
&\Pr\left(X_{n\in[N],j>k}=0\given \Gamma_k\right)\\
=&\Pr\left( \left|\Pi \cap \left(\reals^2\times \nats^N\right)\right| = 0 \given \Gamma_k\right),
\]
which itself can be written explicitly using the fact that the
number of atoms of a Poisson point process in any set has a Poisson distribution:
\[
&\hspace{-.2cm}=\exp\left(-\int_{\gamma \geq \Gamma_k}\sum_{x \in\nats^N}h(x\given \tau(v,\gamma))G(\dee v) \dee\gamma\right)\\
&\hspace{-.2cm}=\exp\left(-\int_{\gamma \geq \Gamma_k}\hspace{-.2cm}\left(1-h\left(0\given \tau(v,\gamma)\right)^N\right)G(\dee v)\dee \gamma\right)\label{eq:error}
\]
If this expression cannot be evaluated exactly but an upper bound is available,
then rejection sampling may be used. It is worth noting that
\cref{eq:error} appears in past bounds on the incurred error by truncating
completely random measures \cite[Eqn 4.2]{campbell19}; here 
we bridge the connection between
truncation and slice sampling; if a tight bound of the truncation
error exists, then efficient rejection sampling scheme can be developed
accordingly. Prior to inference, the integral \cref{eq:error} can be precomputed for a 
range of $\Gamma_k$s; afterward, it can be evaluated via interpolation.

\paragraph{Discard unused traits} Discard $\psi_k$, $V_k$, $\Gamma_k$
for $k > K$; this step will only occur if $K_\text{prev} > K$.

\paragraph{Sample $X$} For each $n\in[N]$ and $k\in[K]$, 
note that $X_{nk}$ is conditionally independent of all other variables
given $V_k, \Gamma_k, U_n, (X_{nj})_{j\neq k}$.
Simulate the integer value of $X_{nk}$ from its full conditional distribution with
measure proportional to
\[
	&f\left(Y_n; X_{nk}\delta_{\psi_k} + \sum_{j\neq k}X_{nj}\delta_{\psi_j}\right)
              h\left(X_{nk}|\tau\left(V_k,\Gamma_k\right)\right)\nonumber\\
	\cdot &\xi\left(\hat{k}_{n}(X_{nk})\right)^{-1}\ind\left[U_n\le\xi\left(\hat{k}_{n}\left(X_{nk}\right)\right)\right],\label{eq:samplex}
\]
where the function $\hat{k}_n : \nats_0 \to \nats_0$ is defined by
\[
\hat{k}_n(x) \defined \left\{ \begin{array}{ll}
                            k'_n & x = 0, k=k_n\\
                             k & x > 0, k > k_n\\
                             k_n & \text{otherwise}
                                \end{array}\right. ,
\] 
i.e., it computes what $k_n$ \emph{would be} as we vary the value of $X_{nk}$.
Note that this step can be parallelized across the $N$ data points.
Further, note that $\left(X_{nk}\right)_{n\in[N],k>K}$ are all guaranteed to be 0
due to the truncation at level $K$, and so do not need to be simulated. If the density 
is intractable, we can use Metropolis-Hasting within Gibbs for this step. 

\paragraph{Update local truncation levels}
For each $n\in[N]$, compute the first and second maximum active indices,
\[
k_n &\gets \max\{k \in \nats : X_{nk} > 0\} \cup \{0\}\\
k'_n &\gets \max\{k < k_n : X_{nk} > 0\} \cup\{0\}.
\]

Most probablistic graphical models with CRM component can be converted to the form of \ref{eq:model}. This algorithm then takes the input of a probablistic graphical model with samplers that sample from the conditional distributions and pointers to the variables $Y$, $X$, and $\psi$. This procedure is shown in pseudocode in \cref{alg:slice}.
\input{series_algorithm}

%% file: graph.tex
\begin{figure}[t]
	\centering
	\begin{tikzpicture}
	
	% Define nodes
	\node[obs]                               (y) {$y_n$};
	\node[latent, above=of y, xshift=-1.2cm]   (x) {$X_{nk}$};
	\node[latent, above=of y, xshift=1.2cm] (psi) {$\psi_k$};
	\node[latent, below=of x, xshift=-1.2cm]   (u) {$U_n$};
	\node[latent, above=of x] (v) {$V_k$};
	\node[latent, above=of x, xshift=1.2cm] (Gamma) {$\Gamma_k$};
	\node[const, above=of v]   (lambda) {$\lambda$};
	\node[const, above=of x, xshift=-1.4cm]   (c) {$\alpha$};
	
	\edge{x,psi}{y}
	\edge{x}{u}
	\edge{v,Gamma,c}{x}
	\edge{lambda}{v}
	
	\plate {n} {(u)(x)(y)} {$N$} ;
	\plate {k} {(v)(Gamma)(x)(psi)} {$\infty$} ;
	
	\end{tikzpicture}
	\caption{Probablistic graphical model based on series representation of CRM in plate notation after augmentation with auxiliary variables $U$. }
	\label{fig:graph}
\end{figure}

%% file: series_algorithm.tex
\begin{algorithm}[t!]
    \caption{Slice sampling for general CRMs}\label{alg:slice}
	\begin{algorithmic}[1]
	\Procedure{Slice Sampler}{$y$, $M$ $K$, $f$, $h$, $\tau$, $G$}
	\State Initialize $X,K_{\text{prev}},K, k,k'$
	\For{$m\gets 1,\cdots,M$}:
		\State Sample $U_n$ for $n\in[N]$
		\State Update $K, K_{\text{prev}}$
		\State Resize $X,\psi,\Gamma,V$ to $K$
		\State Sample $\psi_k$ for $k\in[K]$
		\For{$k\gets 1,\cdots,K$}
			\If{$k < K_\text{prev}$}:
				\State Sample $\Gamma_k,V_k|\Gamma_{k-1},\Gamma_{k+1}$ 
			\Else:
				\State Sample $\Gamma_k,V_k|\Gamma_{k-1}$
			\EndIf
		\EndFor
		\State Sample $X_{nk}$ for $n\in[N]$, $k\in[K]$
		\State Update $k, k'$
	\EndFor
	\EndProcedure
	\end{algorithmic}
\end{algorithm}

%% file: applications.tex
\section{APPLICATIONS}\label{sec:application}
In this section, we show how the general slice sampler from \cref{sec:slicesampling} can be applied to two
popular Bayesian nonparametric models: the beta-Bernoulli latent feature model \cite{hjort90,griffiths05}
and beta-negative binomial (BNB) combinatorial clustering model \cite{broderick15}. In particular, we provide
the details of each of the sampling steps from \cref{sec:gibbssampling} based on
the Bondesson series representation \cite{bondesson82} of the beta process. 
The beta process with mass parameter $\alpha$ and shape parameter $\lambda$ has rate
measure 
\[
	\nu(d\theta)=\lambda\alpha\theta^{-1}(1-\theta)^{\lambda-1}d\theta,
\]
and has a Bondesson representation \citep{bondesson82,campbell19} for the rates
\[
\theta_k &= V_k\exp\left(-\Gamma_k/(\lambda\alpha)\right)\\
\left(\Gamma_k\right)_{k=1}^\infty &\dist \distPP(1), \quad V_k \distiid \distBeta(1, \lambda-1),
\]
where $\distPP(1)$ is shorthand for a unit-rate homogeneous Poisson process on $[0, \infty)$.
The beta process can be paired with a Bernoulli likelihood
\[
	h(x|\theta)=\theta^x\left(1-\theta\right)^{1-x}, \quad x\in\{0,1\},
\]
or negative binomial likelihood,
\[
	h(x|\theta,r)=\binom{x+r-1}{x}\left(1-\theta\right)^r\theta^x, \quad x\in\nats\cup\{0\}.
\]
In both of the following examples, we set the monotone sequence $\xi(k)$ to
$\xi(k) = \exp\left(-k / \Delta_\xi\right)$, where $\Delta_\xi > 0$ is 
a hyperparameter to be tuned in each case.

\subsection{BETA-BERNOULLI FEATURE MODEL}\label{sec:betabern}
%In this section, we use Beta-Bernoulli process as prior for a linear Gaussian
%latent factor model. This model can be used in. The model assumes data being
%generated from a Gaussian distribution whose mean is a additive result of
%latent features. The Beta process prior enforces the emission probability of
%the Bernoulli likelihood and therefore results in a the set of features that
%occured to be sparse. In this model, we assume the data $y$ to be
%matrix-normally distributed whose mean is a dot product of Bernoulli matrix $X$
%and a matrix-normally distributed feature matrix $\psi$. The matrix normal
%distributions of $y$ has independent entries with standard-deviation $\sigma$
%and the matrix normal distributions of $y$ has independent entries with
%standard-deviation $\sigma_0$. 
Given a dataset of observations $y_n\in\reals^d$, $n=1,\dots, N$, the beta-Bernoulli latent feature model
aims to uncover a collection of latent features $\psi_k\in\reals^d$ and binary assignments
$X_{nk}\in\{0,1\}$ of data to features responsible for generating the observations:
\[
	\{\Gamma_k\}_{k=1}^\infty&\dist\distPP\left(1\right)\cr
	X_{nk}&\distind\distBern\left(\exp\left(-\frac{\Gamma_k}{c}\right)\right), \quad n\in[N]\cr
	\psi_k&\distiid\distNorm\left(0,\sigma_0^2I\right), \quad k\in \nats\cr
	y_{n}&\distind\distNorm\left(\sum_{k}X_{nk}\psi_k,\sigma^2I\right), \quad n\in[N].
\]
We assume the hyperparameters $\sigma_0^2$, $\sigma^2$, and $c$ are given. We don't need to sample $V$ because here $G=\delta_1$. 

\paragraph{Sample $X$:}

For each $n\in[N]$ and $k\in[K]$, we sample $X_{nk} = r \in \{0, 1\}$
with probability proportional to
\[
	&\distNorm\left(y_{n}; \sum_{j=1, j\neq k}^\infty X_{nj}\psi_{j}+r\psi_{k},\sigma^2I\right)\cdot\cr
	&\distBern\left(r;\exp\left(-\frac{\Gamma_k}{c}\right)\right)\distUnif\left(U_n;0,\xi\left(\hk_n(r)\right)\right),
\]
where $\distNorm(\cdot;\dots)$, $\distBern(\cdot;\dots)$, and $\distUnif(\cdot;\dots)$ are the density functions
of the respective distributions.
One may parallelize this sampling step across $n\in[N]$ due to the introduction of auxiliary variables.

\paragraph{Sample $\psi$:}
Using the conjugacy of the feature prior and data likelihood, 
we sample all the features $(\psi_k)_{k=1}^K$ simultaneously,
\[
Q&= X^TX+\frac{\sigma^2}{\sigma_0^2}I\\
\psi&\dist\distMN\left(\psi;Q^{-1}\left(X^Ty\right),Q^{-1},\sigma^2I\right),
\]
where $\distMN(\cdot; \dots)$ is the density function for the matrix normal distribution \citep{dawid81}.

\paragraph{Sample $\Gamma$ (substep 1):}
The full conditional distribution of $\Gamma_k$ has density proportional to
\[
	& e^{-m_k\Gamma_k/c}\left(1-e^{-\Gamma_k/c}\right)^{N-m_k}\ind\left[\Gamma_{k-1}\leq \Gamma_k\leq \Gamma_{k+1}\right],
\]
where $m_k \defined \sum_{n=1}^N X_{nk}$. Rather than simulating from this density exactly---which would require expensive iterative numerical integration---we 
use Metropolis-Hastings to sample $\Gamma_k$, with proposal
\[
W &\dist \distUnif\left[-\Delta_\Gamma,\Delta_\Gamma\right]\\
\Gamma'_{k} &= W+\max\left(\min\left(\Gamma_k,\Gamma_{k+1}-\Delta_{\Gamma}\right), \Gamma_{k-1}+\Delta_\Gamma\right)\label{eq:rwgam}
\]
for step size $\Delta_\Gamma>0$ and $\Delta_\Gamma=\frac{\Gamma_{k+1}-\Gamma_{k-1}}{n_\Gamma}$ by dividing the interval length into $n_\Gamma$ pieces. 

\paragraph{Sample $\Gamma$ (substep 2):}
The expansion distribution of $\Gamma_k$ has density proportional to
\[
	&e^{-(1+m_k/c)\Gamma_k - I(\Gamma_k)}\left(1-e^{-\Gamma_k/c}\right)^{N-m_k}\!\!\!\!\ind\left[\Gamma_k \geq \Gamma_{k-1}\right]
\]
where $m_k \defined \sum_{n=1}^N X_{nk}$, and
\[
	I\left(\Gamma_k\right) &=\int_{\Gamma_k}^\infty1-\exp\left(-Ne^{-\gamma/c}\right)d\gamma.
\] 
We again use Metropolis-Hastings to sample and $\Gamma_k$. For convenience we set $\Delta_\Gamma=\frac{1}{n_\Gamma}$ for this step in particular. 

Note that $I(\gamma)$ can be precomputed using numerical integration
at a wide range of points prior to slice sampling; here we chose 
to precompute $I(x)$ at 1000 evenly spaced points $e^{-x/c} \in [\epsilon,1]$ for 
$\epsilon=10^{-30}$.
During MCMC, we evaluate $I(\Gamma_k)$ with spline-interpolation.

\subsection{BNB CLUSTERING MODEL}\label{sec:bnb}
Given a collection of documents $d=1, \dots, D$
each containing $N_d\in\nats$ words $y_{dn} \in [W]$, $W\in\nats$,
BNB combinatorial clustering aims to uncover 
latent topics 
$\psi_k \in [0,1]^V$, $\sum_{w}\psi_{kw} = 1$,
and document-specific topic rates $\pi_{dk} > 0$,
$k=1,\dots,\infty$,
via
\[	
        \{\Gamma_k\}_{k=1}^\infty&\dist\distPP\left(1\right)\cr
	V_k&\distiid\distBeta\left(1,\lambda-1\right)\cr
	\theta_{dk}&\distind\distBeta\left(\alpha\lambda V_ke^{-\frac{\Gamma_k}{c}},\lambda\left(1-\alpha V_ke^{-\frac{\Gamma_k}{c}}\right)\right)\cr
        \psi_{k}&\distiid\distDir\left(\beta\right)\cr
        \pi_{dk}&\distind\distGam\left(r,\frac{1-\theta_{dk}}{\theta_{dk}}\right)\cr
	Z_{dn}&\distind\distCat\left(\left(\frac{\pi_{dk}}{\sum_k\pi_{dk}}\right)_{k=1}^\infty\right)\cr
	y_{dn}&\distind\distCat\left(\psi_{z_{dn}}\right).
\]

Here $Z_{dn}$ is the topic indicator of word $n$ in document $d$. We assume the hyperparameters $\lambda>1$, $0<\alpha<1$ and set $c=\lambda\alpha$. $\beta \in \reals_+^V$, and $r>0$ are given.
In this model, note that we sample per-word auxiliary variables:
\[
	U_{dn}&\dist\distUnif\left[0,\xi\left(Z_{dn}\right)\right].
\]

\paragraph{Sample $Z$}

The conditional distribution of $Z_{dn}$ is a categorical thresholded
by the auxiliary variable; in particular, the probability that $Z_{dn} = k$
is proportional to
\[
 \pi_{dk}\psi_{k y_{dn}}\ind\left[U_{dn}\le\xi\left(k\right)\right]/\xi(k).
\]

% slice variab distribution We sample $Z$ instead of directly sample $X$ as in \ref{eq:samplex}. Then we use $z$ to calculate $X$ as: $X_{dk}=\sum_{n=1}^N\ind\{z_{dn}=k\}$.

\paragraph{Sample $\psi$}

We sample the latent features exactly from their full conditional 
Dirichlet distribution via
\[
\psi_k \dist \distDir\left( \left(\beta_v+\sum_{(n,d):z_{dn}=k}\ind\{y_{dn}=w\}\right)_{w=1}^W\right).
\]
The calculation is standard and similar to \cite{blei03}. 

\paragraph{Sample $V,\Gamma$ (substep 1):}

The full conditional distribution of $V_k$, $\Gamma_k$
has density proportional to 
\[
	&\prod_{d=1}^D\distBetaNegBinom\left(X_{dk};r,\alpha\lambda V_{k}e^{-\frac{\Gamma_{k}}{c}},\lambda\left(1-\alpha V_{k}e^{-\frac{\Gamma_{k}}{c}}\right)\right)\cr
	&\cdot\distBeta\left(V_{k};1,\lambda-1\right)\distUnif\left[\Gamma_{k};\Gamma_{k-1},\Gamma_{k+1}\right]
\]
where $X_{dk} = \sum_{n=1}^{N_d} \ind\left[Z_{nd} = k\right]$,
and $\distBetaNegBinom(\cdot; \dots)$, $\distBeta(\cdot;\dots)$, and $\distUnif(\cdot; \dots)$ are the density functions for the beta-negative binomial, beta, and uniform distributions. Here the $X_{dk}$'s are conditionally independent because we do not condition on a fixed number of words in each document $N_d$.

We again use Metropolis-Hastings to sample $V_k$ and $\Gamma_k$. Here $V$ is sampled by a random walk proposal for with hyperparameter $\triangle_V$. 
\[
	E &\dist \distUnif\left[-\Delta_V,\Delta_V\right]\\
	V'_{k} &= E+\max\left(\min\left(V_k,1-\Delta_V\right), \Delta_V\right),
\]
and $\Gamma$ is sampled using the same algorihtm as \ref{eq:rwgam}. 

\paragraph{Sample $V,\Gamma$ (substep 2):}

The expansion distribution of $V_k$, $\Gamma_k$ has density proportional to
\[
	&\exp\left(-I\left(\Gamma_k\right)\right)\cr
	&\prod_{d=1}^D\distBetaNegBinom\left(X_{dk};r,\alpha\lambda V_ke^{-\frac{\Gamma_k}{c}},\lambda\left(1-\alpha V_ke^{-\frac{\Gamma_k}{c}}\right)\right)\cr
	&\distBeta\left(V_k;1,\lambda-1\right)\distExp\left(\Gamma_k-\Gamma_{k-1};1\right),
\]
where the integral expression is given by
\[
	&I\left(\Gamma_k\right)\cr
	&=\int_{\Gamma_k}^\infty\int_0^1\left(1-F\left(v,\gamma\right)\right)\distBeta\left(v;1,\lambda-1\right)dvd\gamma\\
	&F(v,\gamma) =\distBetaNegBinom\left(0;r,\alpha\lambda ve^{-\frac{\gamma}{c}},\lambda\left(1-\alpha ve^{-\frac{\gamma}{c}}\right)\right)^D.
\]
For simplicity the formula presented here is for beta-negative binomial likelihood with homogeneous failure probability but we set them differently in later experiments.
This density can be jointly sampled using Metropolis-Hastings with thresholded
uniform proposals, similarly to the previous substep. Furthermore,
as in substep 2 of the previous example, this integral
can be precomputed for a range of values before sampling.

%% file: experiments.tex
\section{EXPERIMENTS}\label{sec:experiments}
In this section, we compare the performance of our algorithm on the two models
described in \cref{sec:betabern,sec:bnb}. On both synthetic and real datasets
our algorithm out-performs state-of-the-art methods and fixed
truncation.

\subsection{BETA-BERNOULLI FEATURE MODEL}

In the first experiment we generate synthetic data from a truncated version
of the beta-Bernoulli model from \cref{sec:betabern}, with  model parameters
set to $(\sigma,\sigma_0,c)=(0.2,0.5,1)$. We test a number of experimental settings: for each
$N\in\{10000,11000,\cdots,19000,20000\}$, we set the synthetic generating model truncation level
to $K=2\ceil{\log(N)}$ and data dimension $D=2\ceil{\frac{N\log(N)}{N-\log(N)}}$,
such that the features are roughly identifiable from the data. 
For the auxiliary variables in the slice sampler, we set the scale
of the $\xi$ sequence to $\Delta_\xi=1$. The scale is optimized over $\{0.1,0.2,0.3,0.4,...,2.9,3\}$ to maximize effective sample size per second (ESS/s) using simulated data ($N$=1000). We set the Metropolis-Hastings step size to $n_\Gamma = 10$. This chosen from the set ${1, 2, …, 10}$ to result in a MH acceptance rate of $\Gamma$ averaged over iterations to be roughly between 0.2 and 0.9, which is a standard general practice in MCMC methods.
We generate synthetic data and run the proposed slice sampling algorithm for 1,000 iterations
over 10 independent trials, comparing to the state-of-the-art
accelerated collapsed Gibbs sampler by \citet{doshivelez09acc} with $\mcO(N^2)$ runtime per MCMC iteration. 
We perform the comparison by measuring both ESS/s and 
2-norm error of held-out data. The error is evaluated with latent features
from the Monte Carlo samples and combinatorial variable $X$ chosen to
minimize error. $(N_{train},N_{test},K,c,\sigma,\sigma_0,\triangle_\xi)=(300,200,20,2,0.5,0.5)$ where the 
parameters that require tuning are tuned in a procedure similar to the previous experiment. 
To evaluate the ESS/s for both samplers, we compute a test function that returns 1 if the
combinatorial matrix $X$ has an even number of non-zero entries and 0
otherwise, and use the batch mean estimator
\cite{flegal10}.

\begin{figure*}[t]
\begin{subfigure}{0.5\textwidth}
	\includegraphics[scale=0.11]{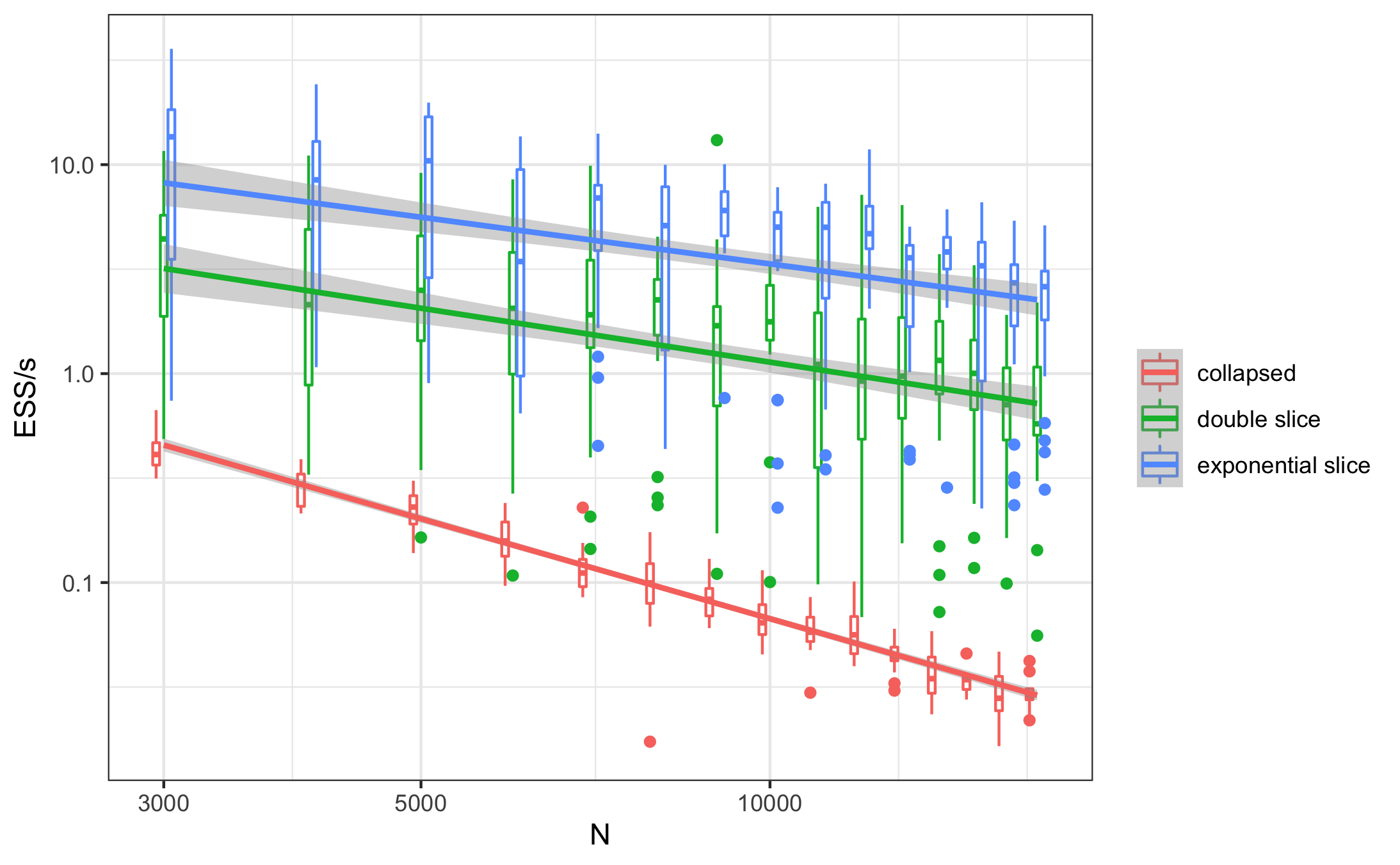}
	\caption{}\label{fig:ess}
\end{subfigure}
\begin{subfigure}{0.5\textwidth}
	\includegraphics[scale=0.11]{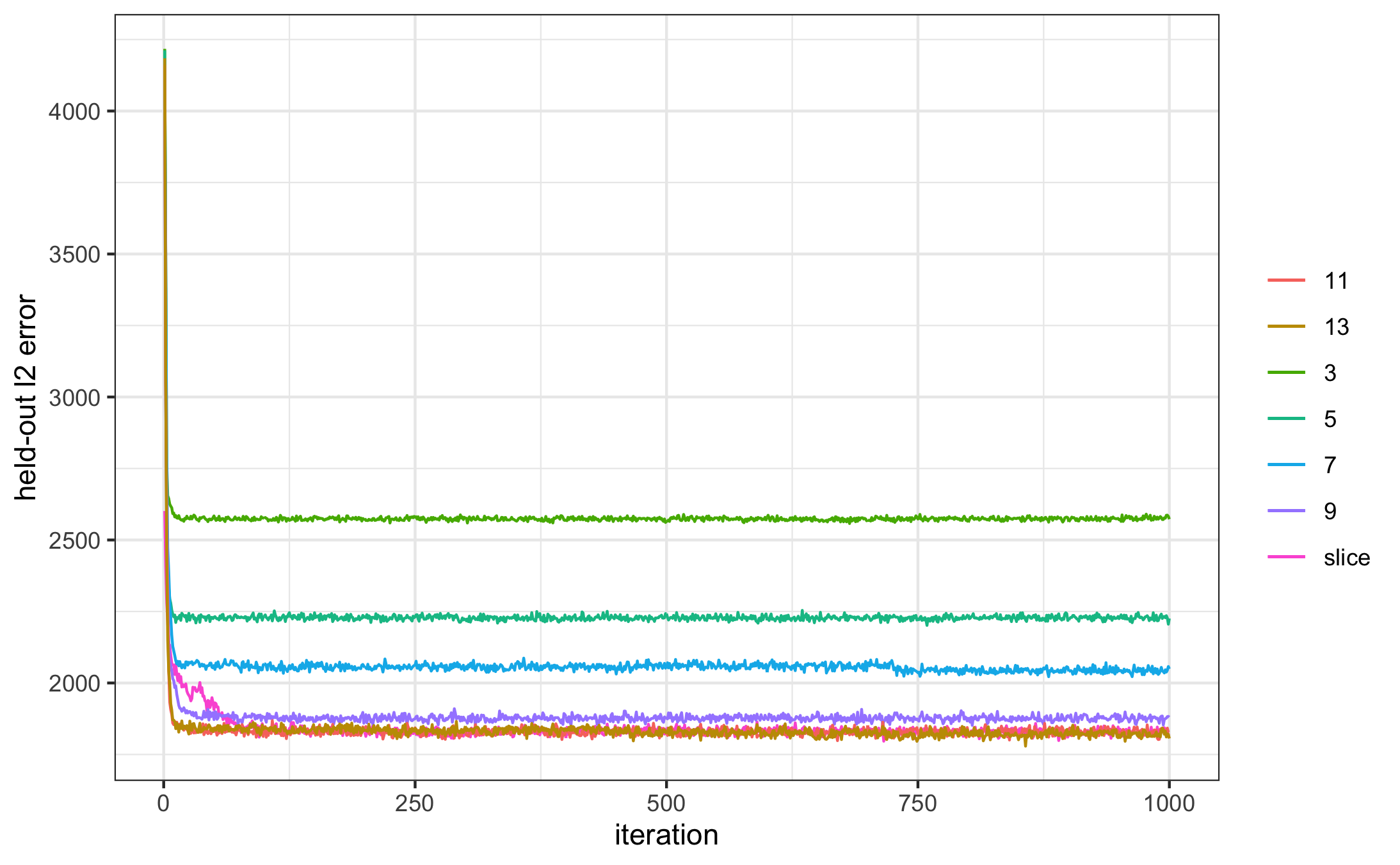}
	\caption{}\label{fig:hll}
\end{subfigure}
\caption{(\ref{fig:ess}): Boxplot and linear fit of the ESS/s of 20
runs of model. Both ESS/s and $N$ are plotted in
log scale of base 10. The line ``double slice'' corresponds to auxiliary
variable $U_n\equiv K_n\dist\distUnif\left[k_n',2k_n'\right]$. 
(\ref{fig:hll}): 2-norm error of held-out data for truncated and adpatively truncated model (slice sampler) optimized over the combinatorial space.}
\end{figure*}

The results are shown in \cref{fig:ess,fig:hll}.
\cref{fig:ess} suggests our sampler has ESS/s that scales as
$\mcO(N^{-0.6})$, while the accelerated collapsed sampler has ESS/s that scales as
$\mcO(N^{-1.34})$; this improvement arises from the linear runtime
per-iteration compared to the quadratic runtime of the algorithm of
\cite{doshivelez09acc}. \cref{fig:hll} achieved the smallest error by quickly selecting a suitable truncation level.

\begin{figure*}[t!]
\begin{subfigure}{0.5\textwidth}
	\includegraphics[scale=0.1]{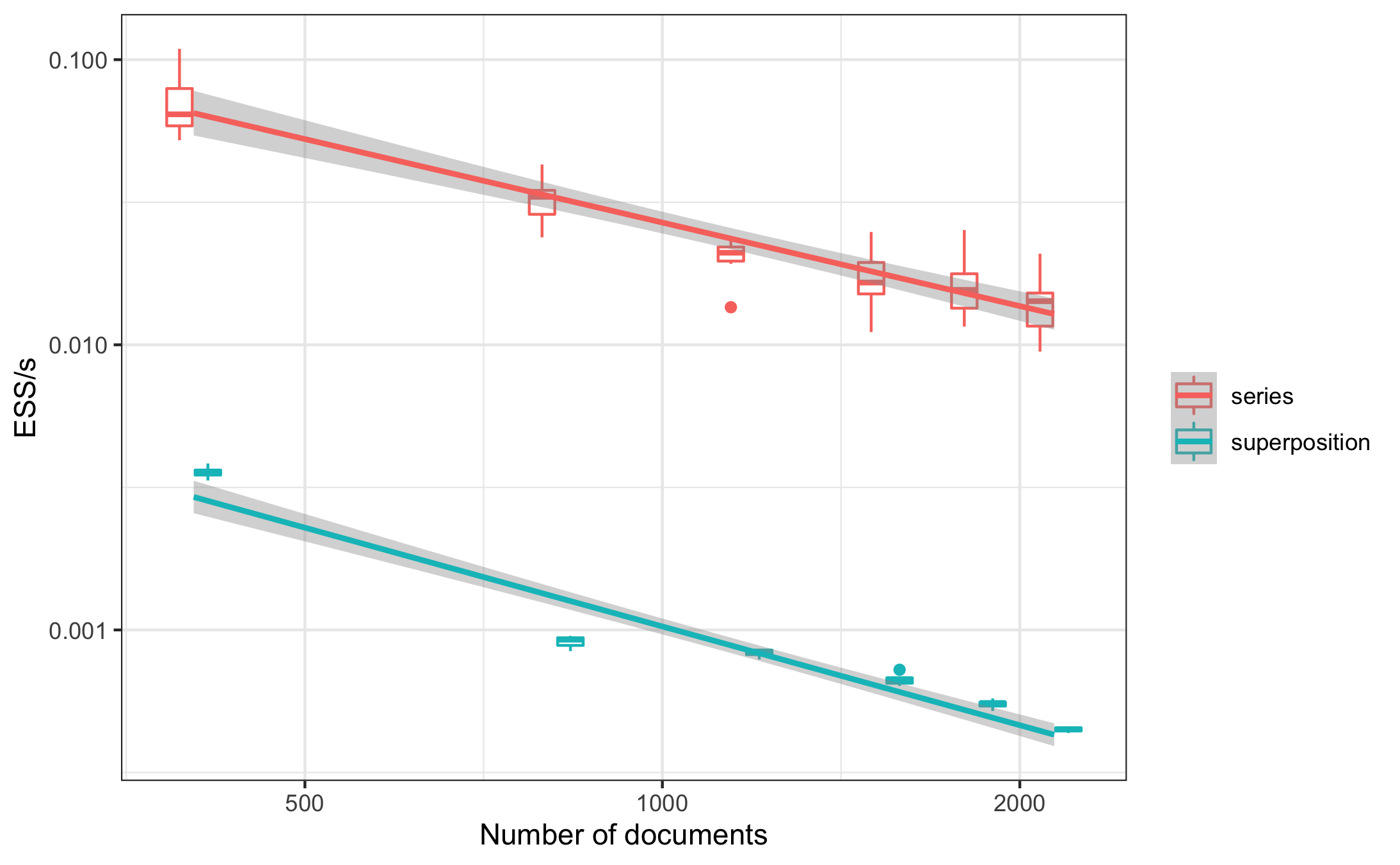}
	\label{fig:essps}
	\caption{}\label{fig:bnbess}
\end{subfigure}
\begin{subfigure}{0.5\textwidth}
	\includegraphics[scale=0.1]{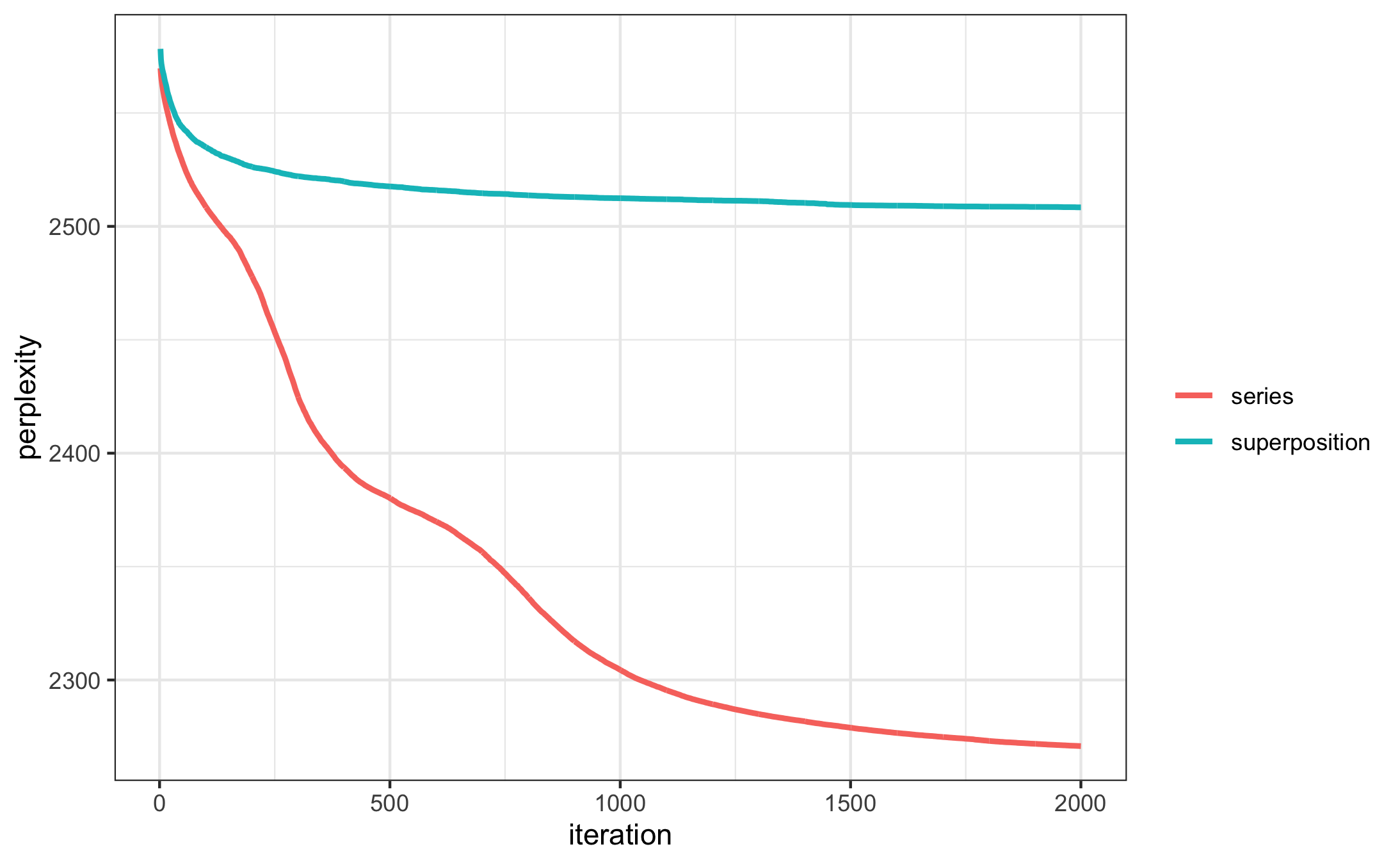}
	\caption{}\label{fig:bnbhll}
\end{subfigure}
\caption{(\ref{fig:bnbess}): Boxplot and linear fit of the 
ESS/s of 6 runs of model. 
Both ESS/s and $N$ are plotted in log scale of base 10. 
(\ref{fig:bnbhll}): Perplexity evaluated on the test set at each iteration. }
\end{figure*}

\subsection{HIERARCHICAL CLUSTERING}

In the second experiment, we used the BNB clustering
model from \cref{sec:bnb} to analyze the NeurIPS corpus
 from 2010 to 2015, preprocessed to remove
stopwords and truncate the vocabulary to those words appearing more
than 50 times.  We randomly split each document of the NeurIPS papers corpus
into held-out test words (30\%) and training words (70\%).
We set concentration and scale parameters to
$(\alpha,\lambda)=(1,1.1)$, the prior Dirichlet topic distribution parameter
 over the  $V$ vocabulary words to
$\beta=(0.1,\cdots,0.1)$, and the failure rate of the negative binomial distribution
is set to $r_d=\frac{N_d(\lambda-1)}{\alpha\lambda}$ for each document $d\in\{1, \dots, D\}$.
We set the Metropolis-Hastings step sizes to $(n_\Gamma, \Delta_V) = (10, 0.3)$,
and the scale of the $\xi$ sequence to $\Delta_\xi = 3$. These parameters are tuned similar to the previous section. 
We compared our slice sampler to the slice sampler of \cite{broderick15}
on both ESS/s and held-out data perplexity. We use the same procedure as in the previous experiment to estimate ESS/s.

The results  are shown in \cref{fig:bnbess,fig:bnbhll}.
\cref{fig:bnbess} shows that our algorithm produces a roughly two orders of magnitude
improvement in ESS/s
on large datasets than the comparison method. \cref{fig:bnbhll} demonstrates
that the proposed slice sampler also provides a significant decrease
in the held-out test set perplexity \cite{blei03}. 
This is at least in part because the proposed slice sampler is generic and can use any
series representation of the underlying CRM; here, we take advantage of that and 
use the Bondesson
representation, which is known to provide exponentially decreasing truncation error \cite{campbell19}
and is significantly more efficient than the superposition
representation used by \cite{broderick15}. In practice, this manifests as a high number of unused or redundant atoms in past samplers, while 
the proposed sampler does not exhibit this  issue.

%% file: conclusion.tex
\section{CONCLUSION}

In this paper, we introduced a computational method for posterior inference in
a large class of unsupervised Bayesian nonparametric models.  Compared with
past work, our method enables parallel inference, does not require conjugacy,
and targets the exact posterior. 

It is worth noting that the proposed sampler does not necessarily generalize 
past slice samplers for specific models (e.g., \cite{teh07}). Although model-specific
methods may provide performance gains in some cases, our method can be
easily incorporated in a general probabilistic programming system, 
and provides parallel inference for a wide range of models.
Future work on the proposed methodology could include
automated selection of the deterministic sequence $\xi$,
and the incorporation of more advanced Markov chain moves,
 such as split-merge \cite{jain07}.

%% file: acknowledgement.tex
\subsubsection*{Acknowledgements}
This research is supported by a National Sciences and Engineering Research Council of Canada (NSERC) Discovery Grant and Discovery Launch Supplement.

%% file: 282_supplementary.tex
\section{APPENDIX}
We now show that the proposed slice sampler defines a valid
Markov chain Monte Carlo algorithm (\cref{thm:supp}). In particular,
(1) the exact posterior $\pi$ is the invariant distribution of the Markov
chain, and 
(2) that a law of large numbers holds: 
for any measurable function $\Phi$ and initial state $S_0$,
the sequence of states $S_1, S_2, \dots$ produced by the slice sampler
satisfies
\[
\frac{1}{T}\sum_{t=1}^T \Phi(S_t) \convas \EE_\pi\left[\Phi(S)\right].
\]

We start with some basic notation. Let $\mcS$ be a set endowed with a $\sigma$-algebra $\mcB$,
and let $\pi$ be a target probability distribution on $\mcS$.
A \emph{Markov kernel} $\kappa : \mcS \times \mcB \to [0, 1]$ satisfies two properties:
(1) for each $B\in\mcB$, $\kappa(\cdot , B) : \mcS \to [0, 1]$ is a measurable function,
and
(2) for each $s\in\mcS$, $\kappa(s, \cdot)$ is a probability measure.
$\kappa(s, B)$ can be thought of as the probability of transitioning 
to any state $s'\in B\subseteq \mcS$ in a single jump starting from a particular state $s\in\mcS$.
Given two Markov kernels $\kappa_1$, $\kappa_2$, define the \emph{composition} $\kappa_1\circ \kappa_2$
of the kernels---another Markov kernel---via
\[
(\kappa_1\circ \kappa_2)(s,B)=\int \kappa_1(s',B)\kappa_2(s,ds').
\]
As with a single kernel, the composition $(\kappa_1\circ \kappa_2)(s, B)$ can be thought of as 
the probability of transitioning to any state $s'\in B\subseteq \mcS$ after two jumps---first
via $\kappa_2$, then via $\kappa_1$---starting from a particular state $s\in\mcS$.

One of the key conditions for a kernel $\kappa$ to create a Markov chain Monte Carlo scheme
for a target distribution $\pi$ is \emph{$\pi$-invariance}: 
if one samples $s\sim \pi$, and then simulates
a transition $s' \sim \kappa(s, \cdot)$, we require that $s' \sim \pi$.
In other words, for any measurable set $B$,
\[
\int \kappa(s,B)\pi(ds)=\pi(B).
\]
We use the following results in \cref{lem:markov} to analyze the $\pi$-invariance of the proposed
slice sampler for the posterior distribution $\pi$.
\bnlem\label{lem:markov}
Let $(\kappa_j)_{j=1}^\infty$ be Markov kernels, and suppose $\mcS$ can be written
as a countable partition $\mcS = \bigcup_j B_j$, $i\neq j \implies B_i \cap
B_j = \emptyset$ of sets of nonzero measure $\pi(B_j) > 0$.
\benum
\item If the $\kappa_j$ are all $\pi$-invariant, and
\[
\kappa(s, B) = \lim_{J\to\infty} \left(\kappa_J \circ \dots \circ \kappa_1\right)(s, B) 
\]
exists pointwise for $s\in\mcS$ and $B\in\mcB$, then
$\kappa$ is a $\pi$-invariant Markov kernel.
\item If each $\kappa_j$ is $\pi_j$-invariant, where 
\[
\pi_j(B) = \frac{\pi(B \cap B_j)}{\pi(B_j)},
\]
then
\[
\kappa(s, B) = \sum_{j=1}^\infty \ind\left[s\in B_j\right] \kappa_j(s, B)
\]
is $\pi$-invariant.
\eenum
\enlem
\bprf
For 1,
\[
&\int \kappa(s,B)\pi(ds)\\
 = &\int \lim_{J\to\infty}\left(\kappa_J \circ \dots \circ \kappa_1\right)(s, B)\pi(ds)\\
 = &\lim_{J\to\infty}\int \left(\kappa_J \circ \dots \circ \kappa_1\right)(s, B)\pi(ds)\\
= & \lim_{J\to\infty} \pi(B) = \pi(B),
\]
where we use the fact that the finite composition of $\pi$-invariant 
kernels is $\pi$-invariant e.g.~by \citep[p.~49]{geyer98}, and Lebesgue
dominated convergence to swap the limit and integral. For 2,
\[
&\int \kappa(s, B)\pi(ds)\\
 =& \sum_{j=1}^\infty \int \ind\left[s\in B_j\right] \kappa_j(s, B) \pi(ds)\\
=& \sum_{j=1}^\infty \pi(B_j) \int \kappa_j(s, B) \frac{\ind\left[s\in B_j\right] \pi(ds)}{\pi(B_j)}\\
=& \sum_{j=1}^\infty \pi(B_j) \pi_j(B)= \sum_{j=1}^\infty \pi(B_j\cap B) = \pi(B),
\]
where we again use Lebesgue dominated convergence to swap the infinite series and integral.
\eprf
Each iteration of the slice sampler
can be written as the kernel composition
\[
\kappa=\Kexp_{\Gamma,V}\circ \kappa_{\Gamma,V}\circ \kappa_X\circ \kappa_\psi\circ \kappa_U.
\]
The kernels $\kappa_{X},\kappa_{\psi},\kappa_{U}$ are the full conditional 
(i.e., Gibbs) kernels for variables $X,\psi,U$;
the kernel $\kappa_{\Gamma, V}$ (substep 1 in the main text) 
is the composition of the full conditional of $\Gamma_k, V_k$ for all $k\in\nats$;
standard results \citep[p.~79]{geyer98} guarantee that each of these is $\pi$-invariant,
and so their composition is $\pi$-invariant by \cref{lem:markov}.
Note that although all of these kernels involve theoretically simulating infinitely many values,
in practice this is unnecessary: truncation by $U$ makes simulating $X_{nk}$ and $\psi_k$
for $k > K$ unnecessary, and we will see that the final kernel $\Kexp_{\Gamma, V}$ overwrites
changes to $\Gamma_k, V_k$ for $k \geq K_\text{prev}$, implying that the full conditional step only 
needs to be run for $k < K_\text{prev}$. 

The only remaining kernel is $\Kexp_{\Gamma, V}$, which corresponds to substep 2 in the main
text. This kernel samples $(\Gamma_k, V_k)_{k=K_\text{prev}}^\infty$ from their full conditional.
Denote $\Kexp_j$ to be the kernel that samples $(\Gamma_k, V_k)_{k=j}^\infty$ from their
full conditional; then
\[
\Kexp_{\Gamma,V} = \sum_{j=0}^\infty \ind\left[K_\text{prev} = j\right] \Kexp_j.
\]
By \cref{lem:markov}, we just need to show that each $\Kexp_j$ is $\pi_j$-invariant,
where $\pi_j$ is the posterior conditioned on $K_\text{prev} = j$, which follows from the fact that $\pi_j$ is a Gibbs kernel. 

We have now shown that the Markov kernel created by the slice sampler
in the main text is $\pi$-invariant. We now complete the final result
in \cref{thm:supp}.
\bnthm\label{thm:supp}
If $f>0$ and $h>0$, then for any measurable function $\Phi$ and 
any initial random state $S_0$, the
sequence of states $S_1, S_2, \dots$ produced by $\kappa$ satisfies
\[
\frac{1}{T}\sum_{t=1}^T\Phi(S_t)\convas\EE_\pi\left[\Phi(S)\right].
\]
\enthm
\bprf
We first establish $\varphi$-irreducibility: let us set $\varphi$  to the posterior distribution, let $s = (v, \gamma, x, \psi, u)$ denote an initial state, and $B$, a target set of configurations with positive posterior probability. 
%Assume without loss of generality that $N=1$ and that $B$ is a rectangle with its side on the $x$ ``axis'' containing a single configuration, i.e. $B = B_v \times B_\gamma \times \{x^*\} \times B_\psi \times B_u$. 
It may not be possible to go from $s$ to $B$ in one application of $\kappa$ as the current configuration of the matrix $x$ constrains what values $u$ can take. However this obstacle disappears by considering paths obtained by two applications of $\kappa$ and visiting an intermediate state where every entry in the matrix $x$ is set to zero. 
To formalize this, let $B_0 = \{(v, \gamma, x, \psi, u) : x_{nk} = 0 \;\forall n, k\}$. Then
\[
\kappa^2(s, B) &= \int \kappa(s, \text{d}s') \kappa(s', B) \\
&\ge \int \mu(\text{d}s') \kappa(s', B)
\]
where $\mu(A) = \kappa(s, A \cap B_0)$. Using the fact that $\xi$ is monotonically decreasing, our assumption that $f$ and $h$ are strictly positive, we obtain from the full conditional of $X$ derived in the paper that $\mu$ is a strictly positive measure on $B_0$. Moreover, using again the same assumptions, straightforward checks on each full conditional derived in the paper shows that provided $s \in B_0$, the function $\kappa(s', B)$ is positive. 

Having established $\varphi$-irreducibility, Harris recurrence follows from \cite[Cor.~13]{roberts_harris_2006} since $\kappa$ is a deterministic alternation of Gibbs kernels. 
Therefore the law of large number follows by \cite[Thm.~17.0.1,~17.1.6]{meyn_markov_1993}.
\eprf